\documentclass[letterpaper, 10 pt, conference]{ieeeconf}  
\IEEEoverridecommandlockouts                              
\usepackage{soul}
\usepackage{url}
\usepackage[utf8]{inputenc}
\usepackage{graphicx}
\usepackage{amsmath}
\usepackage{booktabs}

\usepackage{amssymb,amsfonts}
\usepackage{textcomp}
\usepackage{subfigure}
\usepackage{float}

\usepackage{tabularx}
\usepackage{fancyhdr}
\usepackage{algorithm}    
\usepackage{algorithmic} 
\usepackage[np, autolanguage]{numprint}

\usepackage{epstopdf} %
\usepackage{color,xcolor}
\usepackage{xspace}
\usepackage{balance}

\usepackage{tikz}
\usetikzlibrary{positioning}
\usetikzlibrary{shapes.geometric}
\usetikzlibrary{arrows.meta}
\usetikzlibrary{calc,angles}
\usetikzlibrary[shadings]
\tikzset{>={Stealth[width=3mm]}}
\makeatletter
\newcommand{\gettikzxy}[3]{%
  \tikz@scan@one@point\pgfutil@firstofone#1\relax
  \edef#2{\the\pgf@x}%
  \edef#3{\the\pgf@y}%
}
\newcommand{\markRightAngle}[4][0.2cm]{
    \coordinate (tempa) at ($(#3)!#1!(#2)$);
    \coordinate (tempb) at ($(#3)!#1!(#4)$);
    \coordinate (tempc) at ($(tempa)!0.5!(tempb)$); 
    \draw[line width=1.5pt] (tempa) -- ($(#3)!2!(tempc)$) -- (tempb);
 }
\makeatother

\newcommand{\RefEq}[1]{(\ref{#1})}

\newcommand{\RefFig}[1]{Fig. \ref{#1}}

\newcommand{\et}{$et.al.$}

\usepackage{setspace}

\renewcommand{\baselinestretch}{0.956} 

\title{\LARGE \bf Optimizing Dynamic Balance in a Rat Robot via the Lateral Flexion of a Soft Actuated Spine}

\author{Yuhong Huang$^{1}$,	Zhenshan Bing$^{1,*}$, Zitao Zhang$^{2,3}$, Genghang Zhuang$^{1}$, Kai Huang$^{2,3}$, and Alois Knoll$^{1}$
	\thanks{$^{1}$Authors from the Technical University of Munich, Munich, German
		{\tt\small yuhong.huang@tum.de}}%
	\thanks{$^{2}$Authors from the Sun Yat-Sen University, Guangdong China}%
	\thanks{$^{3}$Authors from the Pazhou Lab, Guangzhou, 510330, China}%
 	\thanks{$^{*}$Corresponding author: Zhenshan Bing}%
}

\begin{document}

\maketitle
\thispagestyle{empty}
\pagestyle{empty}

\begin{abstract}
Balancing oneself using the spine is a physiological alignment of the body posture in the most efficient manner by the muscular forces for mammals. For this reason, we can see many disabled quadruped animals can still stand or walk even with three limbs. This paper investigates the optimization of dynamic balance during trot gait based on the spatial relationship between the center of mass (CoM) and support area influenced by spinal flexion. During trotting, the robot balance is significantly influenced by the distance of the CoM to the support area formed by diagonal footholds. In this context, lateral spinal flexion, which is able to modify the position of footholds, holds promise for optimizing balance during trotting.   This paper explores this phenomenon using a rat robot equipped with a soft actuated spine. Based on the lateral flexion of the spine, we establish a kinematic model to quantify the impact of spinal flexion on robot balance during trot gait. Subsequently, we develop an optimized controller for spinal flexion, designed to enhance balance without altering the leg locomotion. The effectiveness of our proposed controller is evaluated through extensive simulations and physical experiments conducted on a rat robot.  Compared to both a non-spine based trot gait controller and a trot gait controller with lateral spinal flexion, our proposed optimized controller effectively improves the dynamic balance of the robot and retains the desired locomotion during trotting.

\end{abstract}

\section{Introduction}
\begin{figure}[!t]
	\centering
	\input{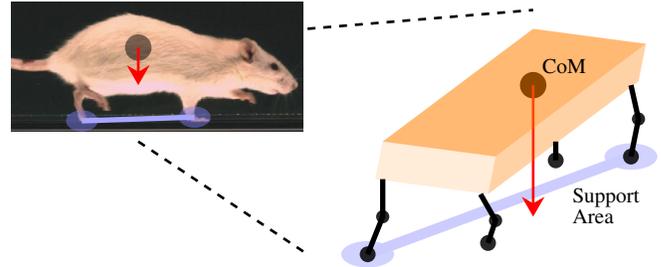}
	\vspace{-1.5em} 
	\caption{The quadruped robot tilts during trot gait. The figure of the trotting rat is cited from \cite{ham2019automated}. The purple line represents the support area during trot gait, while the black point indicates the CoM of the robot. Notably, the projection of the CoM always falls outside the support area.}
  \vspace{-0.75em} 
	\label{fig:case}
\end{figure}

In recent decades,  the capabilities of robots for performing various and complex tasks have been widely explored \cite{gong2010review, xiao2017snake, bing2020perception, bing2021toward, bing2022solving, zhou2023language, wang2023meta, bing2021complex}. Especially, quadruped robots have caught attention for their impressive adaptability on challenging and uneven terrains by choosing suitable footholds \cite{winkler2015planning, biswal2021development, bing2022meta, zhang2023hierarchical, bing2023meta}. This foothold selection also changes the robot's support area dynamically, affecting its balance. Particularly in the context of a walking trot gait, the robot is limited to only two footholds throughout its gait stride.
As shown in \RefFig{fig:case}, this constraint will only lead to a diagonal formed by the leg pairs that can support the robot, which is incapable of continuously covering the center of mass (CoM) of the robot.
When the CoM deviates from the support area,  the robot is prone to losing balance, potentially leading to tilting or even a fall. 
Therefore, how to maintain dynamic balance during trot gaits becomes a great challenge in the control of quadruped robots \cite{ugurlu2013actively, jia2018stability, he2019survey}.

In recent years, extensive research efforts have been dedicated to advancing the control of balance in quadruped robots \cite{ugurlu2013actively, bledt2018cheetah, bing2023lateral}. 
To maintain static balance, the concept of the zero moment point (ZMP) has been proposed \cite{vukobratovic2004zero}. It is utilized to compute CoM trajectory and assess the stability of quadruped robots \cite{ ugurlu2013dynamic, chen2020virtual}. By employing ZMP-based methods, the support area of a robot can be adjusted to cover its CoM, thereby maintaining the balance.
Extending beyond static balance, studies have introduced models for robust dynamic balance to enhance robot locomotion \cite{diedam2008online, mastalli2022agile, sun2022balance}. For instance, Chen \et ~have introduced the dynamic balance tube concept to formalize balance control in quadrupedal locomotion \cite{chen2023quadruped}. Their approach models the quadrupedal locomotion based on a fixed-time gait pattern using switched systems to adjust the relationship between the CoM and support area. 
Many prior research efforts have centered on maintaining the robot's dynamic balance during locomotion by adjusting the foot trajectory and ensuring that the CoM remains within the support area. 
However, these approaches introduce challenges in maintaining the robot's original gait pattern. The robot's foot trajectory may be altered, thereby disrupting the expected locomotion.
As a result, preserving the robot's gait while sustaining dynamic balance poses a significant challenge in the field of quadruped robot control.

In nature, animals can easily maintain balance while walking on a single-plank bridge by swinging their spine left and right without altering their foot trajectory. 
Inspired by this, employing spinal flexion is a potential approach for optimizing balance during robot gait \cite{bing2023lateral}.
The spine and limb locomotion are controlled independently but can work collaboratively \cite{huang2023smooth}. Researchers have enhanced robot locomotion by modifying the stride length through spinal telescoping or flexion, all without involving a change in the robot's gait pattern \cite{chen2017effect, bhattacharya2019learning, huang2022enhanced}.
Due to the complexity of spinal locomotion, there are relatively fewer studies investigating the influence of spinal flexion on the relationship between CoM and support area.
The task of maintaining robot balance during gait by directly applying spinal flexion presents a substantial challenge in this research domain.

Based on our prior work \cite{huang2022enhanced}, this paper presents an optimized spinal flexion controller to enhance the balance maintenance capability during locomotion. To develop this controller, we utilize a rat robot equipped with a soft actuated spine to investigate the impact of spine-based locomotion on dynamic balance during trotting.

Our main contributions are summarized as follows.
\begin{itemize}
    \item To quantify the robot's balance during trotting, we establish a kinematic model incorporating spinal flexion. Based on the footholds affected by lateral spinal flexion, this kinematic model depicts the relationship between the robot's CoM and support area.
    \item To maintain dynamic balance throughout robot gait without altering limb locomotion, we develop an optimized controller for spinal flexion. This controller operates independently from the limb controller and can directly integrate with the default trot gait.  By utilizing this controller, the robot gains the ability to adjust footholds based on the desired balance state.
    \item Compared to other controllers, the proposed optimized controller based on spinal flexion demonstrates the ability to maintain the robot's dynamic balance and desired locomotion during trot gait.
\end{itemize}

\section{Overview of Spine-based Locomotion}
\begin{figure}[!t]
	\centering
	\input{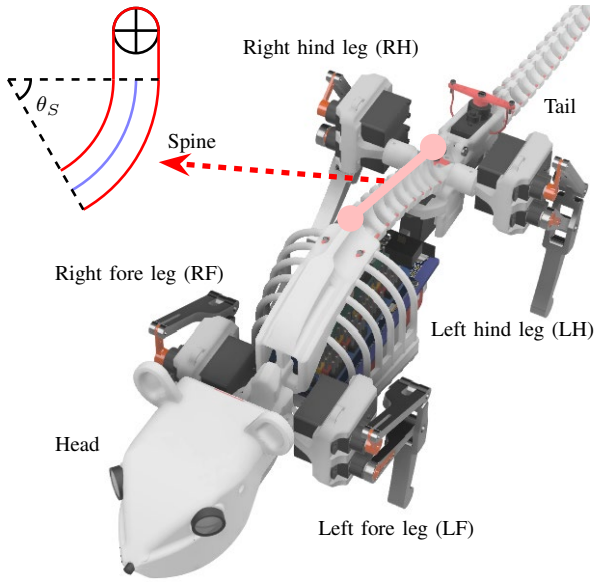}
	\caption{The rat robot with a soft actuated spine. The pink line shows the soft actuated spine. The diagram in the upper right briefly illustrates that the flexing spine can be considered as a segment of a circle with a central angle $\theta_s$.}
 \vspace{-0.5em} 
	\label{fig:robot}
\end{figure}

This section offers an overview of our rat robot, highlighting the unique characteristics of its soft actuated spine.
Furthermore, we briefly present the spine-based locomotion of the rat robot explored in our prior work \cite{huang2022enhanced}.

As shown in \RefFig{fig:robot}, the four limbs of the rat robot are identified as the right fore leg (RF), left fore leg (LF), right hind leg (RH), and left hind leg (LH). 
There are two servos in each leg to stimulate the movement of the hip/shoulder joint and the knee/elbow joint.
Notably, the unique feature of the robot is its soft actuated spine, which is controlled by a tendon-servo system.
Due to its unique structural and material properties, the spine exhibits pure bending behavior, rendering its deformation insignificant when flexed. 
Consequently, the spinal length during spinal flexion is a constant value, denoted as $l_S$. 
Based on the findings from our prior work, the soft actuated spine can be seen as a segment of a circle with a variable central angle $\theta_s$.  And $\theta_s$ depends on the spinal flexion angle $R(t)$, which can be controlled by the linked servos precisely.  This insight enables us to analyze the impact of spinal flexion on hind limb locomotion.

\begin{figure}[!t]
	\centering
	\input{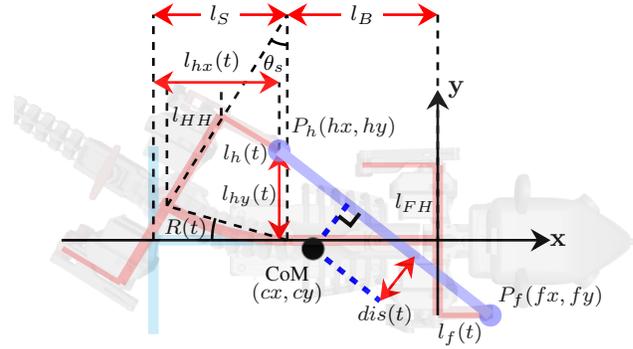}
	\caption{Schematic of the rat robot with spinal flexion. 
	The blue-shaded region represents the robot's skeleton in its initial state, whereas the pink-shaded region depicts the robot's skeleton when flexing its spine.
	$l_{HH}$ and $l_{FH}$ denote the lengths of the hind hip and fore hip, respectively. The dynamic stride lengths of the hind limb and fore limb are $l_h(t)$ and $l_f(t)$ respectively, which generate the robot's gait over time. $l_B$ is the length of the robot body. $l_S$ is the length of the spine. $R(t)$ signifies the time-varying flexing angle of the spine. $P_h$ and $P_f$ represent the footholds for the hind limb and fore limb during trotting, respectively. 
	And the purple-shaded region connecting such two footholds presents the robot's support area. 
	Moreover, to account for the influence of spinal flexion on the hind limb foothold, we introduce $l_{hx}(t)$ and $l_{hy}(t)$ to describe the alterations in the coordinates of the hind foothold due to spinal flexion.}
  \vspace{-0.5em} 
	\label{fig:spine_flexion}
\end{figure}

During the trot gait, the lateral spinal flexion triggers additional displacement at the hips of the hind limbs, resulting in a modified hind limb foothold, as illustrated in \RefFig{fig:spine_flexion}. Within the context of lateral spinal flexion, the extra displacement along the x-axis and y-axis of the hind limb foothold can be expressed as $l_{hx}(t)$ and $l_{hy}(t)$, respectively. Based on the geometric analysis conducted in our prior work \cite{huang2022enhanced}, $l_{hx}(t)$ can be expressed as:
\begin{equation} \label{flexing}
    \begin{split}
    \theta_s = &2R(t), \\
	l_{hx}(t) = &l_h(t)\cos{\theta_s} + l_{HH}\sin{\theta_s} + (l_S - \frac{l_S}{\theta_s}\sin{\theta_s}).
	\end{split}
\end{equation}
The central angle $\theta_s$ of the flexing spine exhibits a linear relationship with the spinal flexion angle $R(t)$. 
$l_{HH}$ and $l_S$ are constant configurations of the rat robot.
$l_f(t)$ and $l_h(t)$ are the control variables of the limb controller that generates robot gait, while $R(t)$ are control variables of the controller controlling spinal flexion.
Therefore, the influence of spinal flexion on footholds can be quantified accurately.

\section{Balance Condition During Trotting}
In this section, we depict the support area of the trot gait according to foothold coordinates.
Subsequently, the robot's balance status is described by the distance from the CoM position to the support area.

\subsection{The Support Area Influenced by Spinal Flexion}
 
To describe footholds in coordinates, we build a local coordinate system for the robot, as illustrated in \RefFig{fig:spine_flexion}. 
The x-axis and y-axis are oriented perpendicular and parallel to the linkage of the robot's shoulders, respectively.
The origin of this coordinate system is positioned at the center of the robot's shoulders.

During trotting, the robot constantly has only two footholds, shown in \RefFig{fig:case}, which we denote as $P_f(fx, fy)$ and $P_h(hx, hy)$. Considering the schematic in \RefFig{fig:spine_flexion},  $fy$ and $fx$ equal to $(-l_{FH})$  and $l_f(t)$, respectively. Additionally, $hx$ is calculated as $(l_{hx}(t) - l_B - l_S)$, while $hy$ corresponds to $l_{hy}(t)$.  Consequently, the expressions for the footholds of the limbs are as follows:
\begin{equation} \label{footholds}
    \begin{split}
    P_f = &(l_f(t), \ -l_{FH}), \\
	P_h = &(l_{hx}(t)-l_B-l_S, \ l_{hy}(t)).
	\end{split}
\end{equation}
Referring to the geometric analysis for $l_{hx}(t)$ conducted in our prior work \cite{huang2022enhanced}, $l_{hy}(t)$ can be expressed as $\frac{l_S}{\theta_s}(1-\cos{\theta_s}) + l_{HH}\cos{\theta_s}-l_h(t)\sin{\theta_s}$. Combined with \RefEq{flexing}, it has
\begin{equation} \label{coordinate}
    \begin{split}
    l_{hx}(t)= &l_h(t)\cos{\theta_s} + l_{HH}\sin{\theta_s} + (l_S - \frac{l_S}{\theta_s}\sin{\theta_s}),\\
    l_{hy}(t) = &\frac{l_S}{\theta_s}(1-\cos{\theta_s}) + l_{HH}\cos{\theta_s}-l_h(t)\sin{\theta_s},\\
	\theta_s = &2R(t).
	\end{split}
\end{equation}
Therefore, the coordinates of $P_h$ can be accurately defined by considering the constant configuration of the rat robot and the given control variables in real time.

As the trot gait always maintains only two footholds, its support area is a line segment formed by two legs in the diagonal. In this context, the support area influenced by spinal flexion during trotting can be defined as a line $P_fP_h$ connecting $P_f(fx, fy)$ and $P_h(hx, hy)$. 
Based on the equation of a straight line, the line $P_fP_h$ can be expressed as $\frac{x - fx}{hx - fx} = \frac{y - fy}{hy - fy}$. 
In other words, $P_fP_h$ is defined as $(hy - fy)x + (fx - hx)y + hx \cdot fy - fx \cdot hy = 0$. 
Combined with \RefEq{footholds}, the support area during trotting with spinal flexion is as follows:
\begin{equation} \label{plane}
    \begin{split}
    (l_{hy}(t)&+l_{FH}) \cdot x + (l_f(t)+l_B+l_S-l_{hx}(t)) \cdot y + \\
	 &(l_B+l_S-l_{hx}(t)) \cdot l_{FH}-l_f(t) \cdot l_{hy}(t) = 0.
	\end{split}
\end{equation}

\subsection{Analysis of Balance Status}
In order to maintain a balanced state during trotting, the position $(cx,cy)$ of the CoM should be above the support area that is denoted as $P_hP_f$.
In other words, the distance $dis(t)$ from the CoM to $P_hP_f$ should be zero at a given time $t$. Thus, the robot's balance status during trot gait can be quantified based on $dis(t)$.
Considering \RefEq{plane}, \RefFig{fig:spine_flexion}, and the formula of distance of a point from a line, 
\begin{equation} \label{dis}
    \begin{split}
    dis(t) =& \frac{A \cdot cx + B \cdot cy + C}{\sqrt{A^2+B^2}} \\
    A = &l_{hy}(t)+l_{FH},\\
    B = &l_f(t)+l_B+l_S-l_{hx}(t),\\
    C = &(l_B+l_S-l_{hx}(t)) \cdot l_{FH}-l_f(t) \cdot l_{hy}(t).
	\end{split}
\end{equation}

To maintain balance status consistently, $dis(t)$ should always remain at zero. 
However, this condition is challenging to be satisfied throughout trotting. 
The support area during trotting is presented as a dynamic diagonal that cannot always cover the CoM. In other words, it is difficult for the robot to maintain balance consistently during trotting. 
In this context, the dynamic balance during trotting can be indicated as the stability of locomotion in
a time slice 
rather than the balance in real-time.
Specifically, we specialize in finding a distribution of balanced status to generate the most stable locomotion in a gait stride.
We assume that the robot will achieve its most stable locomotion during the gait if the balance status occurs at $t = t_b$. In this case, $\lim_{t \to t_b} dis(t) \to 0$. 
The problem of finding such balanced status can be formulated as
\begin{equation} \label{rt}
    \lim_{t \to t_b} dis(t) \to 0 \vdash R(t_b).
\end{equation}
Equation  \RefEq{coordinate} and \RefEq{dis} demonstrate that the value of $dis(t_b)$ can be represented as a functional relationship composed of control variables $l_f(t_b)$, $l_h(t_b)$, and $R(t_b)$. 
Since the trot gait is particularly designed for limb locomotion, $l_f(t_b)$ and $l_h(t_b)$ are given by the gait.
And the remaining unknown control variable is the spinal flexion angle $R(t_b)$.  In this context, $l_f(t_b)$ and $l_h(t_b)$ can be treated as constants, simplifying $dis(t_b)$ to a one-variable equation dependent on $R(t_b)$. Consequently, to achieve $\lim_{t \to t_b} dis(t) \to 0$ , the solution lies in the determination of $R(t_b)$ based on predetermined $l_f(t_b)$ and $l_h(t_b)$.
An analysis of $\frac{\partial dis(t_b)}{\partial R(t_b)}$ readily reveals that $dis(t_b)$ exhibits monotonic behavior within the range $R(t_b) \in \left[\frac{-\pi}{2},\frac{\pi}{2}\right]$. As a result, there exists a unique value of $R(t_b)$ that supports the condition $\lim_{t \to t_b} dis(t) \to 0$, and this value can be efficiently determined using numerical analysis methods.

In summary, for trot gait, there always exists an optimal spinal flexion value that corresponds to a specific distribution of balance status during a gait stride. This distribution can be effectively employed to generate stable locomotion and enhance dynamic balance during a gait stride.

\section{Balance Optimization with Spinal Flexion}
In this section, we explore an optimal distribution of balance status during trotting. 
Following this, we develop an optimized spinal flexion controller, which will operate independently of the limb controller.

\subsection{The Optimal Distribution of Balance Status}
\begin{figure}[!t]
	\centering
	\input{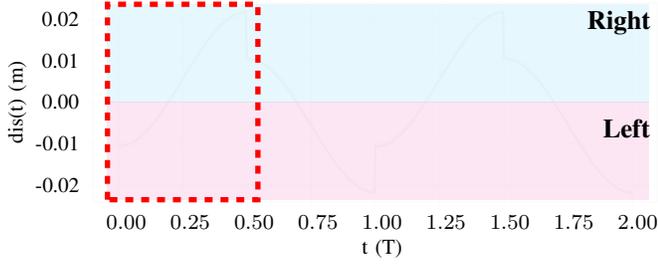}
	\vspace{-1.5em} 
	\caption{The distance from CoM to the support area during trotting without spinal flexion. Considering a gait stride period denoted as $T$, $dis(t)$ is calculated by \RefEq{dis}. Referring to \RefFig{fig:spine_flexion}, when $dis(t) > 0$, CoM is on the left of $P_fP_h$ in the coordinate system, which corresponds to the right side of the support area, as indicated by the blue-shaded region.  Conversely, when $dis(t) < 0$, CoM is on the left of the support area, illustrated by the pink-shaded area. The red dashed rectangle shows the results of a half period.}
	\label{fig:dis_area}
\end{figure}
To investigate the distribution of balance status during trotting, it is necessary to analyze the CoM's distribution over time, which has a significant influence on robot dynamic balance. 
\RefFig{fig:dis_area} indicates the CoM's distribution relative to the robot's support area. 
Within the blue-shaded area, the CoM is situated to the right of the support area, leading to a continuous tilting of the robot to the right. The pink area exhibits an opposite scenario. 
As depicted in \RefFig{fig:dis_area}, the extent of the robot's tilt towards each side is different within a half-period. 
Notably, at the midpoint of a period, the transition between the limb stance phase and swing phase occurs. Specifically, during $t \in [0, \frac{T}{2})$, the robot's footholds are LF and RH. While $t \in [\frac{T}{2}, T)$, the robot's footholds switch to RF and LH.
Consequently, the robot tends to tilt to one side after a stance phase. For instance, when $t=\frac{T}{2}$, the robot tilts to the right. This phenomenon leads to an unbalanced status in the robot's gait, impacting the locomotion of the following stance phase.

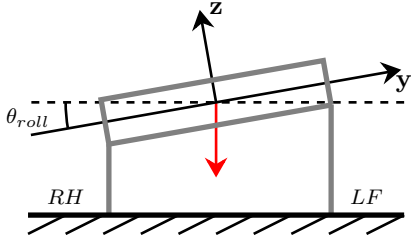
\begin{figure}[!t]
	\centering
	\begin{tikzpicture}
\coordinate (com) at (0,0);
\coordinate (com_0) at (0,-1);
\coordinate (com_1) at (0,0.5);
\coordinate (com_2) at (0,-0.5);

\coordinate (base0_0) at (2.5,0);
\coordinate (base0_1) at (-2.5,0);
\coordinate (base0_2) at (-2,0);
\coordinate (base1_0) at ($(com)!1!10:(base0_0)$);
\coordinate (base2_0) at ($(com)!-1!10:(base0_0)$);
\coordinate (base3_0) at ($(com)!0.5!90:(base1_0)$);
\coordinate (base1) at ($(com)!0.6!10:(base0_0)$);
\coordinate (base2) at ($(com)!-0.6!10:(base0_0)$);

\coordinate (p1) at ($(base1)!0.1!90:(base2)$);
\coordinate (p2) at ($(base1)!-0.1!90:(base2)$);
\coordinate (p3) at ($(base2)!0.1!90:(base1)$);
\coordinate (p4) at ($(base2)!-0.1!90:(base1)$);

\gettikzxy{(p1)}{\pxr}{\pyr};
\gettikzxy{(p4)}{\pxl}{\pyl};
\coordinate (f1) at (\pxr,-1.5);
\coordinate (f4) at (\pxl,-1.5);

\coordinate (g0_1) at (-2.5,-1.5);
\coordinate (g0_2) at (2.5,-1.5);

\draw[color=red,line width=1pt,->](com) -- (com_0);
\draw[color=black,line width=1pt,dashed] (base0_0) -- (base0_1);
\draw[color=black,line width=1pt,<-] (base1_0) -- (base2_0);
\draw[color=black,line width=1pt,->] (com) -- (base3_0);
\draw[color=black,line width=1pt] (base2) -- (p3);
\draw[color=black,line width=1pt] (base2) -- (p4);
\draw[color=gray,line width=2pt] (p1) -- (p2) -- (p3) -- (p4) -- (p1);
\draw[color=gray,line width=2pt] (p1) -- (f1);
\draw[color=gray,line width=2pt] (p4) -- (f4);

\draw[color=black,line width=1pt] (base0_2) arc(-180:-170:2);
\draw[color=black,line width=2pt] (g0_1) -- (g0_2);
\foreach \i in {0, 1, ..., 9}
{
    \draw[color=black,line width=1pt] (-2+\i*0.5,-1.5) -- (-2.5+\i*0.5,-1.75);
}

\node at (-2.5,-0.2) {\footnotesize $\theta_{roll}$};
\node at (0,1.25) {$\mathbf{z}$};
\node at (2.5,0.25) {$\mathbf{y}$};

\node at (-2,-1.25) {\footnotesize $RH$};
\node at (2,-1.25) {\footnotesize $LF$};

\end{tikzpicture}
	\setlength{\abovecaptionskip}{-1pt}
	\caption{Frontal view of the unbalanced robot. $RH$ and $LF$ are two limbs within a stance phase. The red arrow is the direction of gravity. The y-axis and z-axis are based on the local coordinate system of the robot. And $\theta_{\text{roll}}$ is the roll angle of the robot and can indicate the degree of tilt of the robot.}
	\label{fig:body}
	\vspace{-0.5em}
\end{figure}

In \RefFig{fig:body}, the robot's status at $t=\frac{T}{2}$ is depicted. As the robot tilts to the right with a roll angle $\theta_{\text{roll}}$, the limb $RH$ will experience a shorter swing phase, causing the robot to tilt backward. When transitioning footholds to the $RF$ and $LH$ at the same time, $RH$ will still be in the stance phase, starting its swing phase later. This leads to an unbalanced robot status after a stance phase, introducing partial errors to the robot's gait over time. 
To prevent adverse effects on robot locomotion following a stance phase, $\theta_{\text{roll}}$ should be zero at $t=\frac{T}{2}$, thereby allowing the limbs to correctly switch between swing and stance phases.
To achieve this requirement, it is essential to ensure that the extent of the robot's tilt towards each side, influenced by the duration of the tilted state, remains evenly distributed throughout a stance phase. 
Consequently, the balance state of the trot gait should be strategically synchronized with the midpoint of a desired stance phase,
specifically at $dis(\frac{(2n+1)T}{4}) = 0, n \in \mathbb{N}$. In this context, $t_b$ in \RefEq{rt} should be set as $t_b = \frac{(2n+1)T}{4}$.
And the distribution of balance status for optimizing the robot's trot gait can be expressed as:
\begin{equation} \label{balance_status}
   \textstyle \lim_{t \to \frac{(2n+1)T}{4}} dis(t) \to 0, n \in \mathbb{N}.
\end{equation}

\subsection{Spinal Flexion for Balance Optimization}
\begin{figure}[!t]
	\centering
	\begin{tikzpicture}
\coordinate (s0) at (-2,0);
\coordinate (s1) at (2,0);

\draw [gray,opacity=1,line width=2pt] (s0) ellipse [x radius=1.25, y radius=0.75];
\draw [gray,opacity=1,line width=2pt] (s1) ellipse [x radius=1.25, y radius=0.75];

\draw[color=black,line width=2pt,->] (-1,0.5) arc(110:70:3);
\draw[color=black,line width=2pt,->] (1,-0.5) arc(-70:-110:3);

\node at (-2.75,0.25) {$k=$};
\node at (-2,-0.25) {\footnotesize $\frac{2\arccos{\frac{R'}{\alpha}}}{\pi}$};
\node at (1.25,0.25) {$k=$};
\node at (2,-0.25) {\footnotesize $2-\frac{2\arccos{\frac{R'}{\alpha}}}{\pi}$};

\node at (0,1) {\footnotesize $t=\frac{(2n+1)T}{4}$};
\node at (0,-1) {\footnotesize $t=\frac{nT}{2}$};

\end{tikzpicture}
	\setlength{\abovecaptionskip}{-1pt}
	\caption{The binary state machine for changing the scale value $k$. The $\alpha$ is the amplitude of the spinal flexion. $R'$ is the spinal flexion value that maintains the robot with a balance status.}
  \vspace{-0.5em} 
	\label{fig:state}
\end{figure}
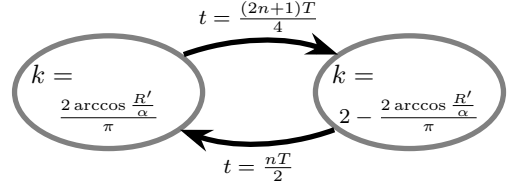

Combining \RefEq{rt} and \RefEq{balance_status}, we can always find a constant value $R(\frac{(2n+1)T}{4})$ that ensures $\lim_{t \to \frac{(2n+1)T}{4}} dis(t) \to 0$.
For simplicity,we define $|R(\frac{(2n+1)T}{4})| = R'$. In our previous work \cite{huang2022enhanced}, the controller for spinal flexion is defined as $R(t) = \alpha\cos{(\omega t +\varphi)}$, where $\alpha$ represents the amplitude of spinal flexion, and $\varphi$ denotes the initial phase. Let $R(t') = \alpha\cos{(\omega t'+\varphi)} = R'$. 
This implies that $t' = \frac{\arccos{\frac{R'}{\alpha}}-\varphi}{\omega}$, which is not always equal to $\frac{(2n+1)T}{4}-\varphi$. To ensure $\lim_{t \to \frac{(2n+1)T}{4}} dis(t) \to 0$, the timeline values need to be scaled using a scaling factor $k$.
Since $\varphi$ does not influence the scaling processes in this context, we can assume $\varphi$ to be zero in the following calculations. 
Let's take the first stance phase as an example, where $t \in [0, \frac{T}{2})$. 
The scaling processes on the timeline can be represented as mapping  $[0,t')$ to $[0, \frac{T}{4})$ and mapping $[t',\frac{T}{2})$ to $[\frac{T}{4},\frac{T}{2})$.
To map $[0,t')$ to $[0, \frac{T}{4})$, $\frac{kT}{4}=t'$. As $\omega = \frac{2\pi}{T}$, $k=\frac{2\arccos{\frac{R'}{\alpha}}}{\pi}$. 
Similarly, when $t \in [\frac{T}{4},\frac{T}{2})$, $k = 2-\frac{2\arccos{\frac{R'}{\alpha}}}{\pi}$.
Based on this, the change in the scale factor $k$ can be expressed as a binary state machine, as shown in \RefFig{fig:state}.

The controller for controlling spinal flexion over time can be effectively designed using a discrete model based on the scale value $k$ with a minimum time step denoted as $t_s$. The scaled value on the timeline can be expressed as $f_T(t) = f_T(t-t_s) + k\omega = f_T(t-t_s) + k\frac{2\pi}{T}$.  Thus, the spinal controller for balance optimization is
\begin{equation} \label{spine_value}
    \begin{split}
    R(t) = &\alpha\cos{(f_T(t)+\varphi)},\\
    f_T(t) = &f_T(t-t_s) + k\frac{2\pi}{T}.
	\end{split}
\end{equation}
Considering \RefEq{spine_value}, \RefEq{balance_status} always holds true. Specifically, when the robot transitions between the stance and swing phases of a limb during trotting, $\theta_{\text{roll}} =  0$. This enables the robot to optimize its gait balance over time, thereby mitigating locomotion errors caused by imbalance.

\section{Experiments}
This section presents the simulation and real-world experiments designed to evaluate the dynamic balance capability of the rat robot during trotting. 
We compare the performance of three distinct controllers, thereby demonstrating the effectiveness of the proposed controller in enhancing dynamic balance during trotting.

\subsection{Experiment Setup}
\begin{figure}[!t]
	\centering
	\input{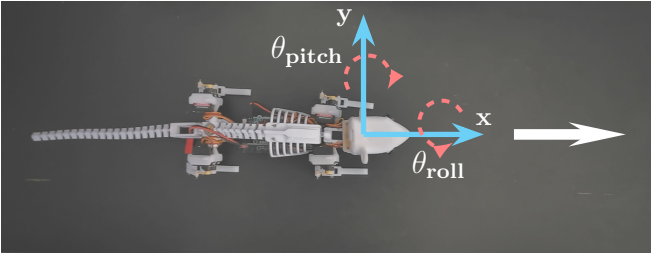}
	\setlength{\abovecaptionskip}{-6pt}
	\caption{Experiment setup used in both physical experiments and simulations. The task requires the robot to walk straight along its x-axis, as the white arrow shows. The roll angle $\theta_{\text{roll}}$  characterizes the lateral tilt of the robot, while the pitch angle $\theta_{\text{pitch}}$ depicts the forward and backward tilting action of the robot. During experiments, $\theta_{\text{roll}}$ and $\theta_{\text{pitch}}$ are applied to evaluate the dynamic balance of the robot.} 
	\label{fig:environment}
\end{figure}

The experimental setup is depicted in \RefFig{fig:environment}. The robot is controlled to go straight along the x-axis. 
The roll and pitch angles are measured via the IMU sensor to evaluate the dynamic balance during trot gait. 
To compare the dynamic balance of different controllers, the robot is controlled with stride frequencies determined as $0.5 + 0.4m$, where $m \in [0,10]$ and $m \in \mathbb{N}$.
For each distinct stride frequency, we execute ten sets of repeated experiments, each starting from a random initial state.
A video accompanying the paper shows both simulation and real-world experiment behaviors.

This paper examines three controllers in the comparative experiments: a default trot gait controller, a trot gait controller incorporating spinal flexion proposed in our prior work \cite{huang2022enhanced} (referred to as ``non-spine'' and ``spine'' controllers, respectively), and the novel ``balance-spine'' controller specifically designed to optimize robot balance through spinal flexion. 
Among them, the ``balance-spine'' controller is composed of the default trot gait controller that directly adds the proposed spinal controller.
As illustrated in \RefFig{fig:body}, the dynamic balance of the robot is quantified as the robot's tilt angle after completing half of a gait stride. 
In this context, the mean values of roll angle and pitch angle during a half stride period are consistently employed across all experiment results as performance indicators, denoted as $\theta_{\text{roll}}$ and $\theta_{\text{pitch}}$, respectively.

\subsection{Influence on Locomotion}
\begin{figure}[!t]
	\centering
	\input{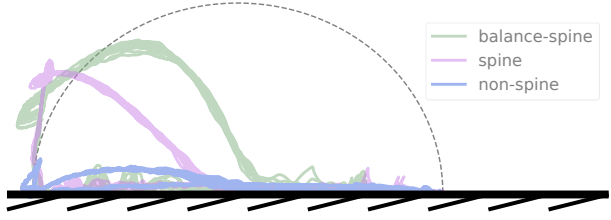}
	\caption{The trajectory of a hind limb walking on the ground using different controllers. The dashed black line represents the initial trajectory configuration for the limb.  This figure intuitively displays the stance phase and swing phase of a limb. When the trajectory overlaps the ground, the limb is a foothold and in the stance phase.}
  \vspace{-0.5em} 
	\label{fig:foot_path}
\end{figure}

\begin{figure}[!t]
	\centering
	\includegraphics[width=.9\columnwidth]{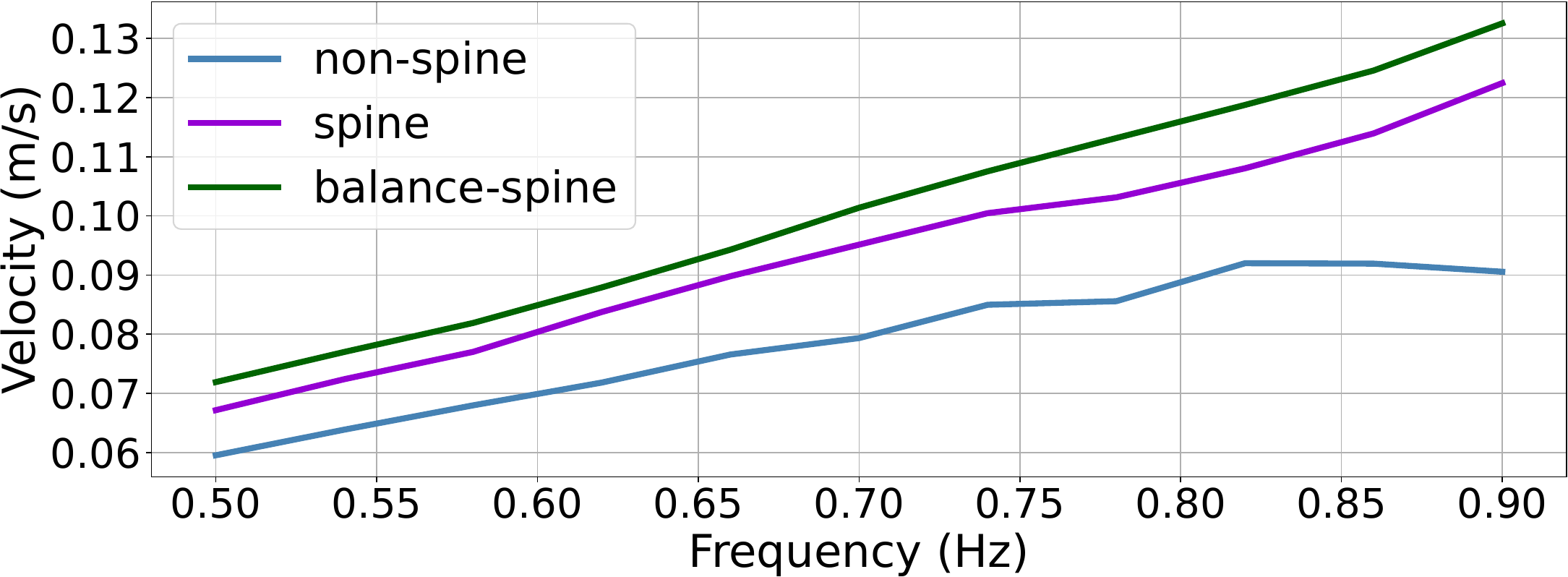}
	\caption{The velocity of the robot controlled by different controllers.}
	\label{fig:vel}
\end{figure}

Analyzing the findings presented in \RefFig{fig:body}, an unbalanced condition may lead to the robot tilting backward, consequently causing a late start of the hind limb's swing phase. 
For instance,  the trajectory of one of the hind limbs is depicted in \RefFig{fig:foot_path}. 
When employing the default trot gait controller, the hind limb exhibits minimal time touching the ground, leading to significant deviations from the desired locomotion.
This phenomenon explains that unexpected behaviors happen during trotting without balance optimization. In other words, an improved dynamic balance enables enhanced robot locomotion. \RefFig{fig:foot_path} clearly illustrates that the foot trajectory under the ``balance-spine'' controller exhibits a more extended swing phase compared to the ``spine'' controller.
This observation indicates that the robot, under the guidance of the proposed controller, maintains a more stable stance posture and achieves better balance during trotting. 
This characteristic translates into significantly increased running speed, as depicted in \RefFig{fig:vel}.
Across all experiments during diverse stride frequencies, the robot controlled by the proposed controller consistently achieves the highest running speeds.
Furthermore, in comparison to the ``spine'' and ``non-spine", the proposed controller can enhance the velocity of the robot by up to 109.0\% and 147.8\%, respectively.
In summary, the proposed controller effectively preserves the predefined limb actions and enhances the robot's locomotion.

\begin{figure}[!t]
	\centering
	\subfigure[The mean of $\theta_{\text{pitch}}$ during trotting in repeated experiments,]{
		\includegraphics[width=.95\columnwidth]{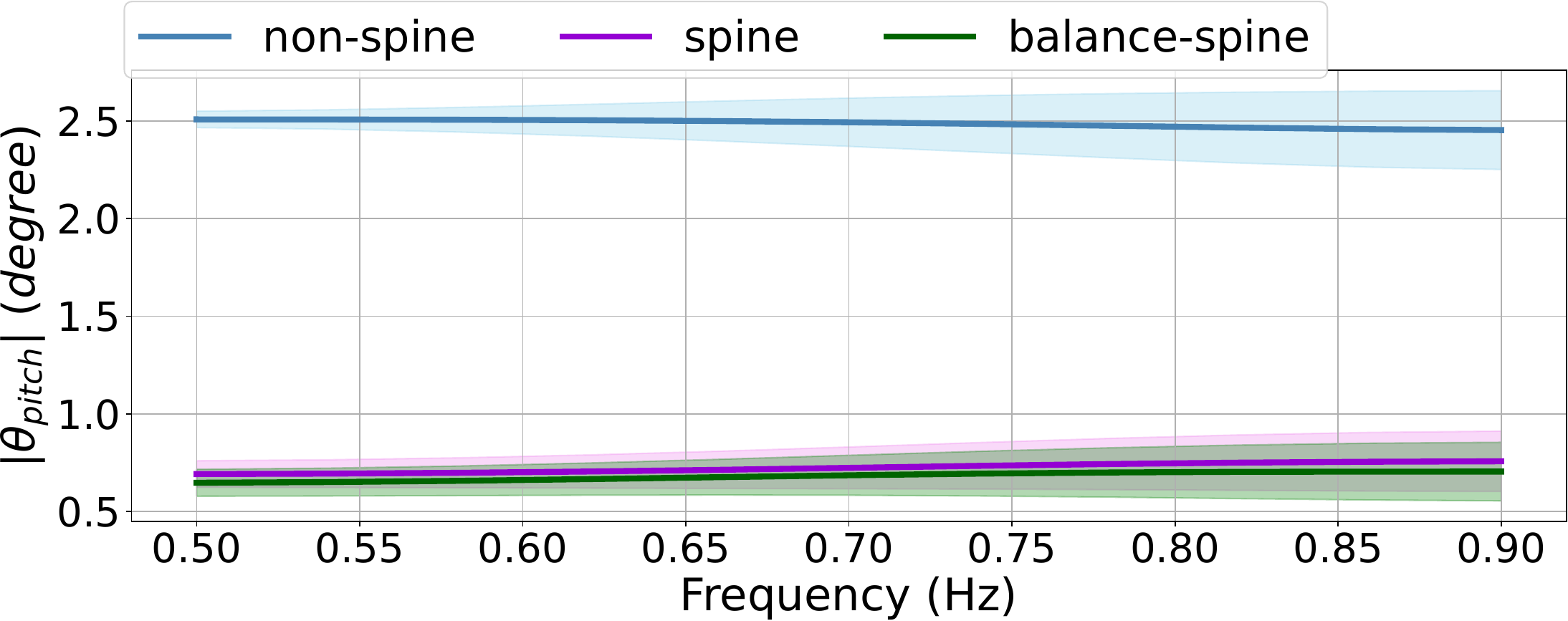}
		\label{fig:sim_pitch}
	}
	\subfigure[The mean of $\theta_{\text{roll}}$ during trotting in repeated experiments,]{
		\includegraphics[width=.95\columnwidth]{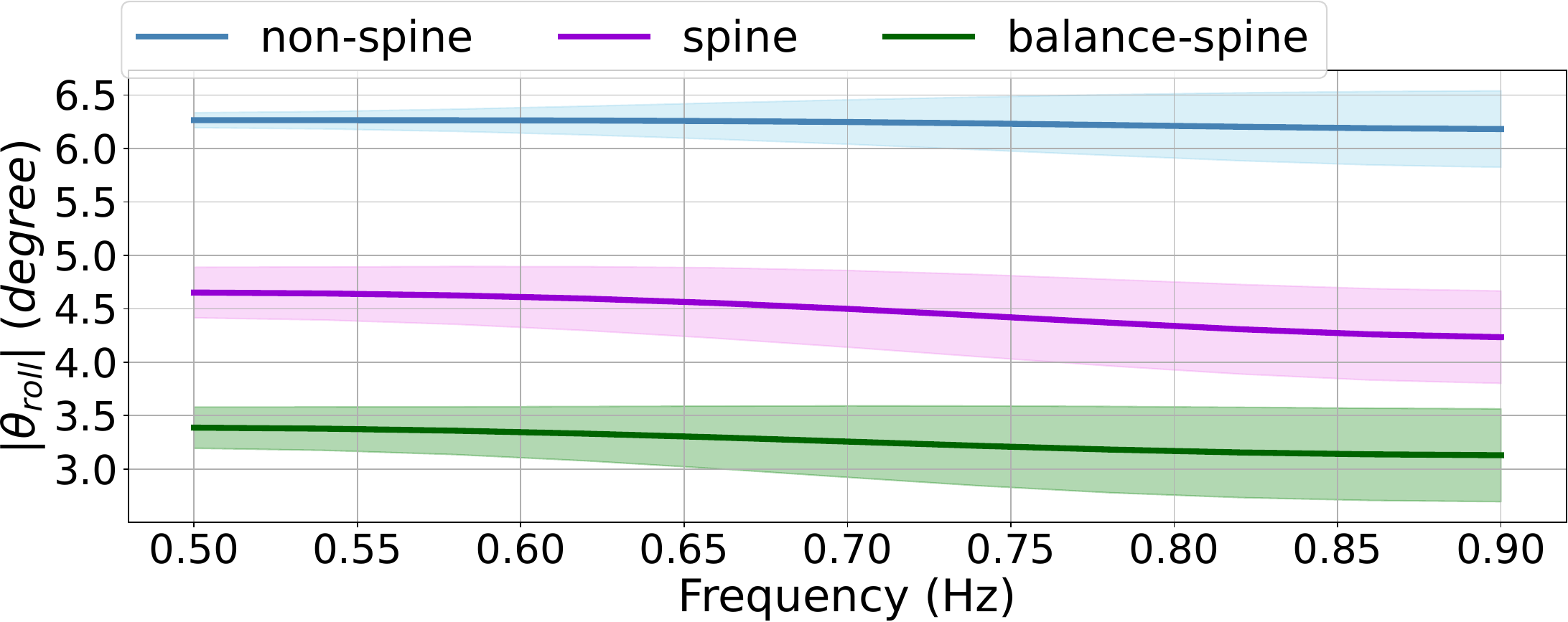}
		\label{fig:sim_roll}
	}
	\caption{Performances to indicate the dynamic balance. The ``non-spine'', ``spine'', and ``balance-spine'' are different controllers with definitions in the experiment setup. The shaded areas present the results of repeated experiments done with random initial robot status.}
  \vspace{-0.5em} 
	\label{fig:angles}
\end{figure}

\begin{figure*}[!bth]
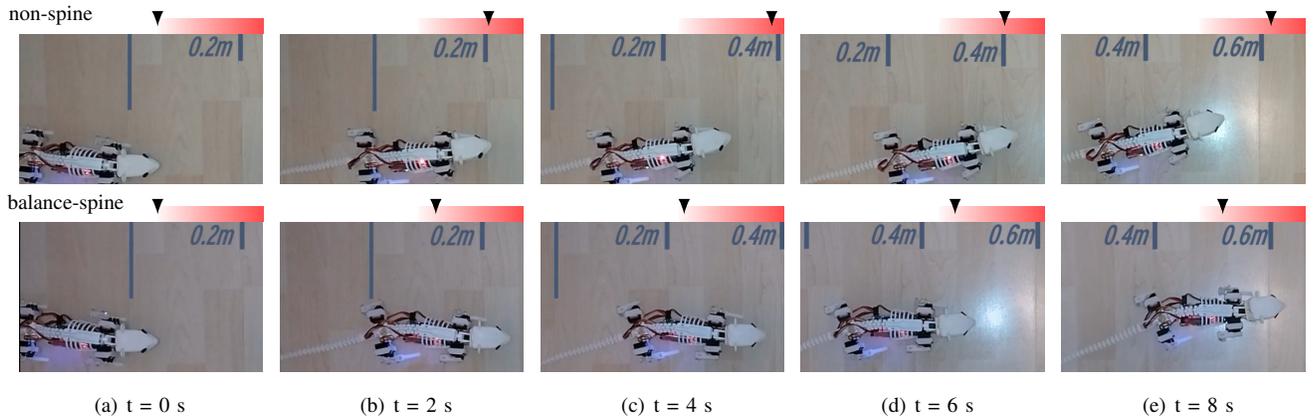

	\centering
	\subfigure[t = 0 s]{
	    \input{Figs/exp/time/tikz/real_s0}
	}\hspace{-5mm}
	\subfigure[t = 2 s]{
		\input{Figs/exp/time/tikz/real_s2}
	}\hspace{-5mm}
	\subfigure[t = 4 s]{
		\input{Figs/exp/time/tikz/real_s4}
	}\hspace{-5mm}
	\subfigure[t = 6 s]{
		\input{Figs/exp/time/tikz/real_s6}
	}\hspace{-5mm}
	\subfigure[t = 8 s]{
		\input{Figs/exp/time/tikz/real_s8}
	}\hspace{-5mm}
 \vspace{-0.5em} 
	\caption{Montage of the trotting gaits in the real world. The first row shows the movement of the rat robot controlled by ``non-spine''. The second row shows the movement of the rat robot controlled by ``balance-spine''. The color bar positioned at the upper right corner of each sub-figure depicts $\theta_{\text{roll}} \in [0^\circ ,8^\circ]$. And the real-time $\theta_{\text{roll}}$ are presented as the arrows on the color bar. Notably, the darker color refers to the larger angle.}
  \vspace{-0.75em} 
	\label{fig:walkStraight}
\end{figure*}

\subsection{Balance Analysis in Simulations}
To explore the dynamic balance of robots controlled by various controllers,  $\theta_{\text{roll}}$ and $\theta_{\text{pitch}}$, which indicates the robot's balance, are analyzed across different stride frequencies during trotting. 
In \RefFig{fig:sim_pitch}, the pitch angle of the robot controlled by the ``non-spine'' significantly exceeds that of other controllers. 
This larger pitch angle consistently causes the robot controlled by the ``non-spine'' to tilt backward during trotting, as previously discussed, resulting in a shortened hind limb swing phase. 
Moreover, the ``balance-spine'' controller exhibits slightly superior performance compared to the ``spine'' controller and controls the robot with pitch angles near zero. In this context, the robot guided by spinal flexion exhibits an insignificant pitch angle and successfully preserves its predefined limb actions and desired locomotion during trotting.
In \RefFig{fig:sim_roll}, $\theta_{\text{roll}}$ quantifies the extent of lateral oscillation experienced by the robot during trotting. 
It is evident that the robot under the influence of the proposed controller consistently exhibits minimal oscillation angles across various stride frequencies. 
This finding demonstrates the superior dynamic balance performance and enhanced locomotion stability achieved by the robot when controlled by the proposed controller. Although the proposed controller is unable to maintain the robot in balance status all the time, it can greatly optimize the dynamic balance of the robot's gait and maintain its desired locomotion.

\subsection{Real-World Experiments}
\begin{figure}[!t]
	\centering
	\subfigure[The mean of $\theta_{\text{pitch}}$,]{
		\includegraphics[width=.465\columnwidth]{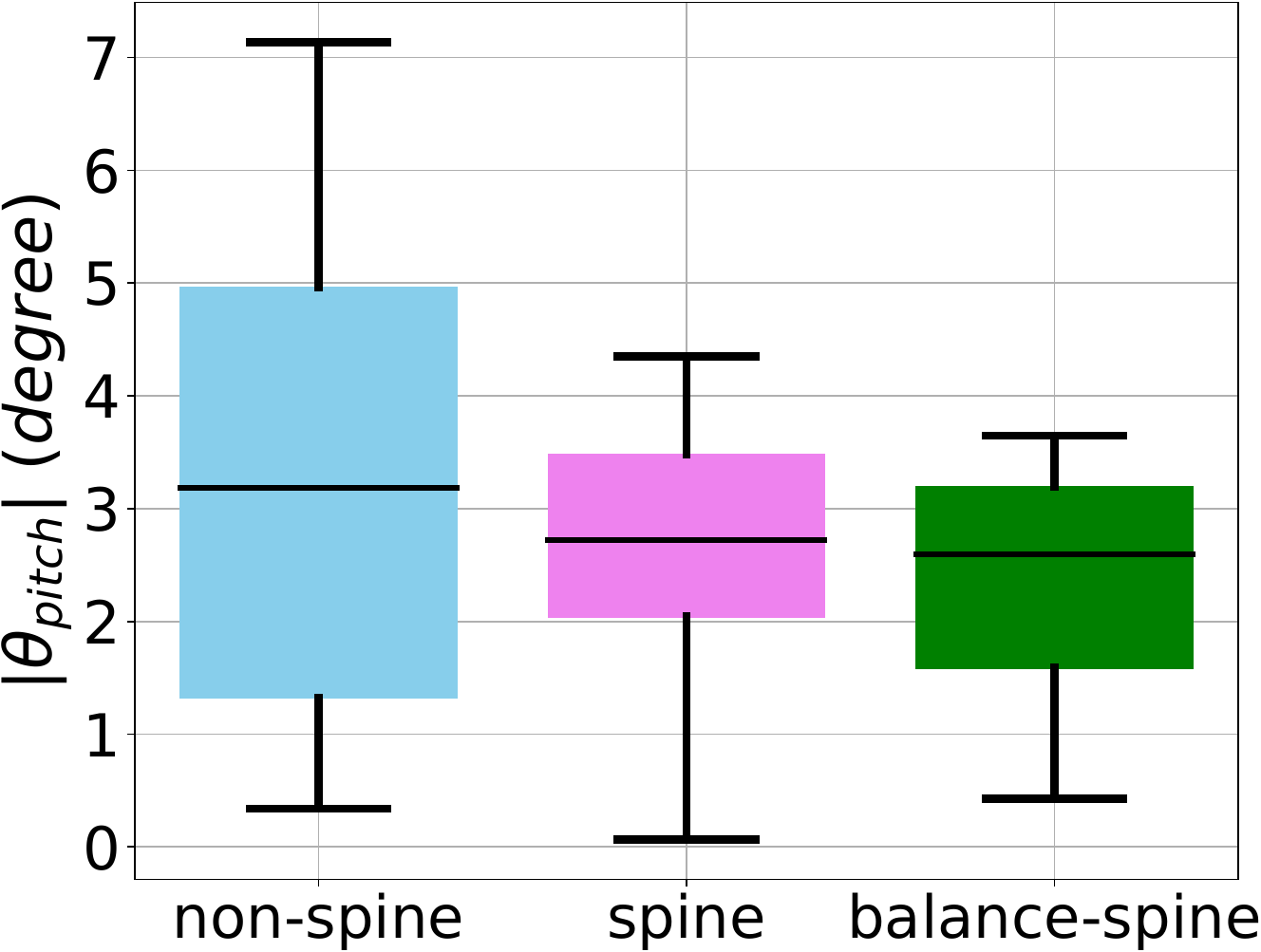}
		\label{fig:real_)pitch}
	}
	\subfigure[The mean of $\theta_{\text{roll}}$, ]{
		\includegraphics[width=.465\columnwidth]{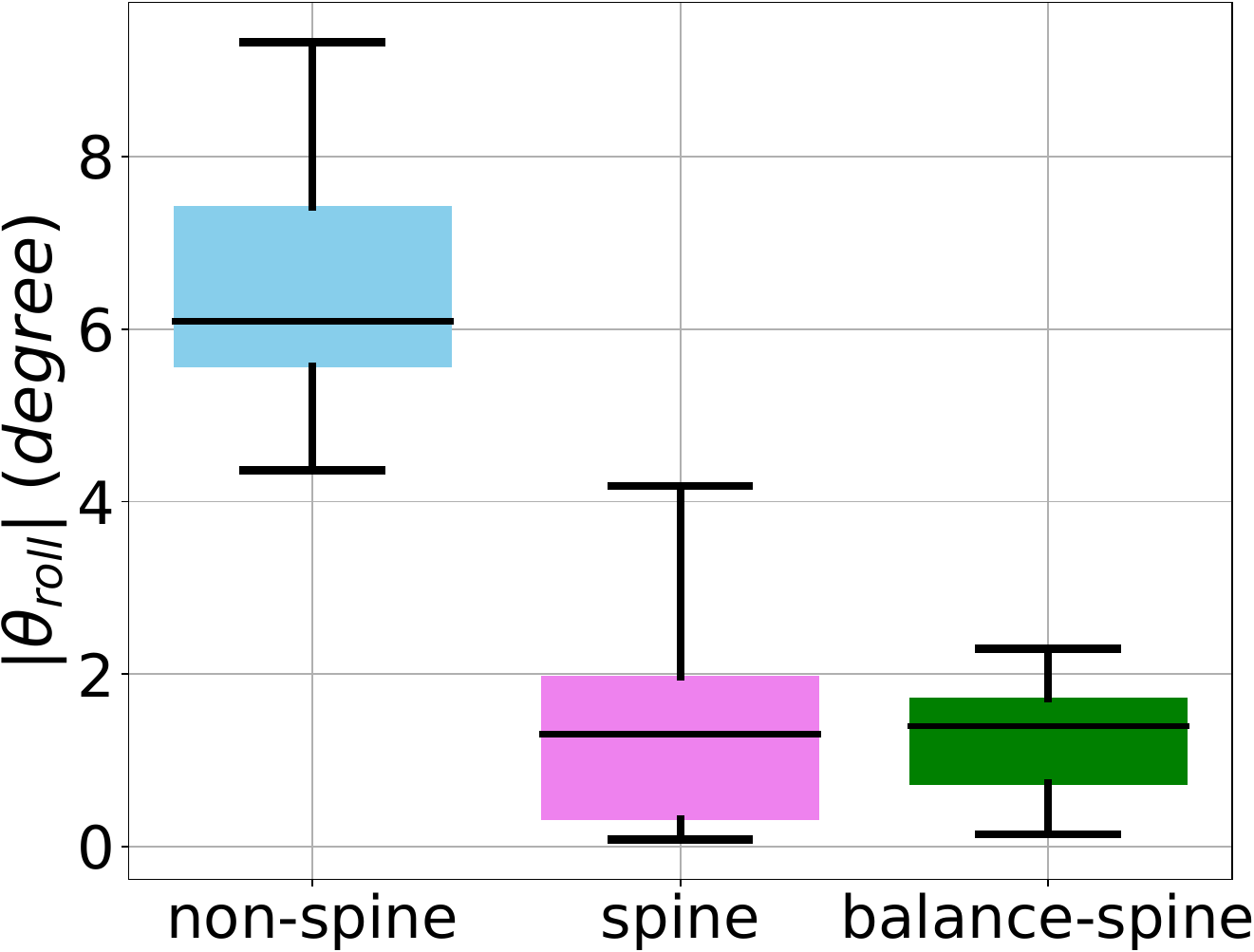}
		\label{fig:real_roll}
	}
	\caption{The performance of dynamic balances of physical robots. Repeated experiments are done based on stride frequencies equal to 0.5 Hz. And the mean value of $\theta_{\text{pitch}}$ and $\theta_{\text{roll}}$ is recorded during trotting.}
  \vspace{-0.75em} 
	\label{fig:real_angle}
\end{figure}

To further validate the effectiveness of our proposed controller, we conducted experiments using a physical rat robot. The montage of the robot walking along its x-axis is shown in \RefFig{fig:walkStraight}. 
At each moment, the color bar on the upper right corner depicts $\theta_{\text{roll}} \in [0^\circ ,8^\circ]$. The darker color corresponds to a larger $\theta_{\text{roll}}$. 
As illustrated in the color bars, $\theta_{\text{roll}}$ of the robot controlled by ``non-spine'' during trotting is always larger than the other.
To further analyze the dynamic balance of different controller, the results of repeated experiments are illustrated in \RefFig{fig:real_angle}. 
In comparison to the simulation results, the observed $\theta_{\text{pitch}}$ during physical experiments exhibits a slight decline in performance. 
This discrepancy may be caused by the natural curvature of the soft spine, which responds to gravity.
Nevertheless, $\theta_{\text{pitch}}$ consistently remains at lower values with minimal fluctuations when utilizing the proposed controller.
When analyzing $\theta_{\text{roll}}$, the proposed controller consistently demonstrates superior performance in terms of maintaining dynamic balance compared to the default trot gait controller. 
The $\theta_{\text{roll}}$ of the proposed controller also exhibits minimal fluctuation. The minimal variation observed in both $\theta_{\text{pitch}}$ and $\theta_{\text{roll}}$ indicates the high level of stability achieved by the proposed controller in controlling robot locomotion.
In summary, our proposed controller exhibits enhanced stability in regulating the robot's locomotion, resulting in an overall improvement in dynamic balance during its gait.

\section{Conclusions}
This paper investigates the optimization of dynamic balance during robot gait based on spinal flexion.
By analyzing the influence of spinal flexion on robot's support area,  we develop a spinal controller to optimize dynamic balance in accordance with the desired balanced status.
This spinal controller is an independent controller decoupled from the limb controller. The experimental results consistently depict the effectiveness of the proposed controller, as evidenced by the robot consistently maintaining minimal pitch and roll angles. In this context, the robot controlled by the proposed controller manifests excellent dynamic balance during the gait.

For future work, we will explore the potential of cooperation of spinal flexion and limb locomotion. This future work will generate a gait that not only runs faster but is also more robust in terms of stability.

\section{Acknowledgment}
This research was funded by the European Union's Horizon 2020 Framework Programme for Research and Innovation under Specific Grant Agreement No. 945539 (Human Brain Project SGA3). Additionally, it received partial support from Pazhou Laboratory Grants (No. PZL2021KF0020).

\balance
\bibliographystyle{IEEEtran}
\bibliography{icra24}

\begin{thebibliography}{10}
\providecommand{\url}[1]{#1}
\csname url@rmstyle\endcsname
\providecommand{\newblock}{\relax}
\providecommand{\bibinfo}[2]{#2}
\providecommand\BIBentrySTDinterwordspacing{\spaceskip=0pt\relax}
\providecommand\BIBentryALTinterwordstretchfactor{4}
\providecommand\BIBentryALTinterwordspacing{\spaceskip=\fontdimen2\font plus
\BIBentryALTinterwordstretchfactor\fontdimen3\font minus
  \fontdimen4\font\relax}
\providecommand\BIBforeignlanguage[2]{{%
\expandafter\ifx\csname l@#1\endcsname\relax
\typeout{** WARNING: IEEEtran.bst: No hyphenation pattern has been}%
\typeout{** loaded for the language `#1'. Using the pattern for}%
\typeout{** the default language instead.}%
\else
\language=\csname l@#1\endcsname
\fi
#2}}

\bibitem{gong2010review}
D.~Gong, J.~Yan, and G.~Zuo, ``A review of gait optimization based on
  evolutionary computation,'' \emph{Applied Computational Intelligence and Soft
  Computing}, vol. 2010, 2010.

\bibitem{xiao2017snake}
S.~Xiao, Z.~Bing, K.~Huang, and Y.~Huang, ``Snake-like robot climbs inside
  different pipes,'' in \emph{2017 IEEE International Conference on Robotics
  and Biomimetics (ROBIO)}.\hskip 1em plus 0.5em minus 0.4em\relax IEEE, 2017,
  pp. 1232--1239.

\bibitem{bing2020perception}
Z.~Bing, C.~Lemke, F.~O. Morin, Z.~Jiang, L.~Cheng, K.~Huang, and A.~Knoll,
  ``Perception-action coupling target tracking control for a snake robot via
  reinforcement learning,'' \emph{Frontiers in Neurorobotics}, vol.~14, p.
  591128, 2020.

\bibitem{bing2021toward}
Z.~Bing, A.~E. Sewisy, G.~Zhuang, F.~Walter, F.~O. Morin, K.~Huang, and
  A.~Knoll, ``Toward cognitive navigation: Design and implementation of a
  biologically inspired head direction cell network,'' \emph{IEEE Transactions
  on Neural Networks and Learning Systems}, vol.~33, no.~5, pp. 2147--2158,
  2021.

\bibitem{bing2022solving}
Z.~Bing, H.~Zhou, R.~Li, X.~Su, F.~O. Morin, K.~Huang, and A.~Knoll, ``Solving
  robotic manipulation with sparse reward reinforcement learning via
  graph-based diversity and proximity,'' \emph{IEEE Transactions on Industrial
  Electronics}, vol.~70, no.~3, pp. 2759--2769, 2022.

\bibitem{zhou2023language}
H.~Zhou, X.~Yao, Y.~Meng, S.~Sun, Z.~Bing, K.~Huang, and A.~Knoll,
  ``Language-conditioned learning for robotic manipulation: A survey,''
  \emph{arXiv preprint arXiv:2312.10807}, 2023.

\bibitem{wang2023meta}
M.~Wang, Z.~Bing, X.~Yao, S.~Wang, H.~Kai, H.~Su, C.~Yang, and A.~Knoll,
  ``Meta-reinforcement learning based on self-supervised task representation
  learning,'' in \emph{Proceedings of the AAAI Conference on Artificial
  Intelligence}, vol.~37, no.~8, 2023, pp. 10\,157--10\,165.

\bibitem{bing2021complex}
Z.~Bing, M.~Brucker, F.~O. Morin, R.~Li, X.~Su, K.~Huang, and A.~Knoll,
  ``Complex robotic manipulation via graph-based hindsight goal generation,''
  \emph{IEEE Transactions on neural networks and learning systems}, vol.~33,
  no.~12, pp. 7863--7876, 2021.

\bibitem{winkler2015planning}
A.~W. Winkler, C.~Mastalli, I.~Havoutis, M.~Focchi, D.~G. Caldwell, and
  C.~Semini, ``Planning and execution of dynamic whole-body locomotion for a
  hydraulic quadruped on challenging terrain,'' in \emph{2015 IEEE
  International Conference on Robotics and Automation (ICRA)}.\hskip 1em plus
  0.5em minus 0.4em\relax IEEE, 2015, pp. 5148--5154.

\bibitem{biswal2021development}
P.~Biswal and P.~K. Mohanty, ``Development of quadruped walking robots: A
  review,'' \emph{Ain Shams Engineering Journal}, vol.~12, no.~2, pp.
  2017--2031, 2021.

\bibitem{bing2022meta}
Z.~Bing, D.~Lerch, K.~Huang, and A.~Knoll, ``Meta-reinforcement learning in
  non-stationary and dynamic environments,'' \emph{IEEE Transactions on Pattern
  Analysis and Machine Intelligence}, vol.~45, no.~3, pp. 3476--3491, 2022.

\bibitem{zhang2023hierarchical}
Z.~Zhang, Y.~Huang, Z.~Zhao, Z.~Bing, A.~Knoll, and K.~Huang, ``A hierarchical
  reinforcement learning approach for adaptive quadruped locomotion of a rat
  robot,'' in \emph{2023 IEEE International Conference on Robotics and
  Biomimetics (ROBIO)}.\hskip 1em plus 0.5em minus 0.4em\relax IEEE, 2023, pp.
  1--6.

\bibitem{bing2023meta}
Z.~Bing, L.~Knak, L.~Cheng, F.~O. Morin, K.~Huang, and A.~Knoll,
  ``Meta-reinforcement learning in nonstationary and nonparametric
  environments,'' \emph{IEEE Transactions on Neural Networks and Learning
  Systems}, 2023.

\bibitem{ugurlu2013actively}
B.~Ugurlu, K.~Kotaka, and T.~Narikiyo, ``Actively-compliant locomotion control
  on rough terrain: Cyclic jumping and trotting experiments on a
  stiff-by-nature quadruped,'' in \emph{2013 IEEE international conference on
  robotics and automation}.\hskip 1em plus 0.5em minus 0.4em\relax IEEE, 2013,
  pp. 3313--3320.

\bibitem{jia2018stability}
Y.~Jia, X.~Luo, B.~Han, G.~Liang, J.~Zhao, and Y.~Zhao, ``Stability criterion
  for dynamic gaits of quadruped robot,'' \emph{Applied Sciences}, vol.~8,
  no.~12, p. 2381, 2018.

\bibitem{he2019survey}
J.~He, J.~Shao, G.~Sun, and X.~Shao, ``Survey of quadruped robots coping
  strategies in complex situations,'' \emph{Electronics}, vol.~8, no.~12, p.
  1414, 2019.

\bibitem{bledt2018cheetah}
G.~Bledt, M.~J. Powell, B.~Katz, J.~Di~Carlo, P.~M. Wensing, and S.~Kim, ``Mit
  cheetah 3: Design and control of a robust, dynamic quadruped robot,'' in
  \emph{2018 IEEE/RSJ International Conference on Intelligent Robots and
  Systems (IROS)}.\hskip 1em plus 0.5em minus 0.4em\relax IEEE, 2018, pp.
  2245--2252.

\bibitem{bing2023lateral}
Z.~Bing, A.~Rohregger, F.~Walter, Y.~Huang, P.~Lucas, F.~O. Morin, K.~Huang,
  and A.~Knoll, ``Lateral flexion of a compliant spine improves motor
  performance in a bioinspired mouse robot,'' \emph{Science Robotics}, vol.~8,
  no.~85, p. eadg7165, 2023.

\bibitem{vukobratovic2004zero}
M.~Vukobratovi{\'c} and B.~Borovac, ``Zero-moment point—thirty five years of
  its life,'' \emph{International journal of humanoid robotics}, vol.~1,
  no.~01, pp. 157--173, 2004.

\bibitem{ugurlu2013dynamic}
B.~Ugurlu, I.~Havoutis, C.~Semini, and D.~G. Caldwell, ``Dynamic trot-walking
  with the hydraulic quadruped robot—hyq: Analytical trajectory generation
  and active compliance control,'' in \emph{2013 IEEE/RSJ International
  Conference on Intelligent Robots and Systems}.\hskip 1em plus 0.5em minus
  0.4em\relax IEEE, 2013, pp. 6044--6051.

\bibitem{chen2020virtual}
G.~Chen, S.~Guo, B.~Hou, and J.~Wang, ``Virtual model control for quadruped
  robots,'' \emph{IEEE Access}, vol.~8, pp. 140\,736--140\,751, 2020.

\bibitem{diedam2008online}
H.~Diedam, D.~Dimitrov, P.-B. Wieber, K.~Mombaur, and M.~Diehl, ``Online
  walking gait generation with adaptive foot positioning through linear model
  predictive control,'' in \emph{2008 IEEE/RSJ International Conference on
  Intelligent Robots and Systems}.\hskip 1em plus 0.5em minus 0.4em\relax IEEE,
  2008, pp. 1121--1126.

\bibitem{mastalli2022agile}
C.~Mastalli, W.~Merkt, G.~Xin, J.~Shim, M.~Mistry, I.~Havoutis, and
  S.~Vijayakumar, ``Agile maneuvers in legged robots: a predictive control
  approach,'' \emph{arXiv preprint arXiv:2203.07554}, 2022.

\bibitem{sun2022balance}
W.~Sun, X.~Tian, Y.~Song, B.~Pang, X.~Yuan, and Q.~Xu, ``Balance control of a
  quadruped robot based on foot fall adjustment,'' \emph{Applied Sciences},
  vol.~12, no.~5, p. 2521, 2022.

\bibitem{chen2023quadruped}
H.~Chen, Z.~Hong, S.~Yang, P.~M. Wensing, and W.~Zhang, ``Quadruped
  capturability and push recovery via a switched-systems characterization of
  dynamic balance,'' \emph{IEEE Transactions on Robotics}, 2023.

\bibitem{ham2019automated}
T.~R. Ham, M.~Farrag, A.~M. Soltisz, E.~H. Lakes, K.~D. Allen, and N.~D.
  Leipzig, ``Automated gait analysis detects improvements after intracellular
  $\sigma$ peptide administration in a rat hemisection model of spinal cord
  injury,'' \emph{Annals of biomedical engineering}, vol.~47, pp. 744--753,
  2019.

\bibitem{huang2023smooth}
Y.~Huang, Z.~Bing, Z.~Zhang, K.~Huang, F.~O. Morin, and A.~Knoll, ``Smooth
  stride length change of rat robot with a compliant actuated spine based on
  cpg controller,'' in \emph{2023 IEEE/RSJ International Conference on
  Intelligent Robots and Systems (IROS)}.\hskip 1em plus 0.5em minus
  0.4em\relax IEEE, 2023, pp. 331--338.

\bibitem{chen2017effect}
D.~Chen, N.~Li, H.~Wang, and L.~Chen, ``Effect of flexible spine motion on
  energy efficiency in quadruped running,'' \emph{Journal of Bionic
  Engineering}, vol.~14, no.~4, pp. 716--725, 2017.

\bibitem{bhattacharya2019learning}
S.~Bhattacharya, A.~Singla, D.~Dholakiya, S.~Bhatnagar, B.~Amrutur, A.~Ghosal,
  S.~Kolathaya, \emph{et~al.}, ``Learning active spine behaviors for dynamic
  and efficient locomotion in quadruped robots,'' in \emph{2019 28th IEEE
  International Conference on Robot and Human Interactive Communication
  (RO-MAN)}.\hskip 1em plus 0.5em minus 0.4em\relax IEEE, 2019, pp. 1--6.

\bibitem{huang2022enhanced}
Y.~Huang, Z.~Bing, F.~Walter, A.~Rohregger, Z.~Zhang, K.~Huang, F.~O. Morin,
  and A.~Knoll, ``Enhanced quadruped locomotion of a rat robot based on the
  lateral flexion of a soft actuated spine,'' in \emph{2022 IEEE/RSJ
  International Conference on Intelligent Robots and Systems (IROS)}.\hskip 1em
  plus 0.5em minus 0.4em\relax IEEE, 2022, pp. 2622--2627.

\end{thebibliography}

\end{document}